\documentclass[runningheads]{llncs}
\usepackage[T1]{fontenc}
\usepackage{graphicx,verbatim}
\usepackage{color}
\usepackage{amsmath}
\usepackage{graphicx}
\usepackage{verbatim}
\usepackage{booktabs,multirow,rotating,colortbl,xcolor}
\usepackage{hyperref} 
\usepackage{xcolor} 

\newcommand{\bestcell}{\cellcolor{bestbg}}
\newcommand{\gainval}[1]{{\textcolor[RGB]{0,128,80}{\textbf{\ensuremath{\downarrow}#1}}}}
\newcommand{\loss}[1]{{\textcolor{red}{\textbf{\ensuremath{\uparrow}#1}}}}
\usepackage{ulem}
\usepackage{xurl} 

\usepackage{colortbl}
\usepackage{rotating}
\usepackage{array}       

\definecolor{lightW5}{RGB}{230,230,230}
\definecolor{secbg}{RGB}{220,220,220}
\definecolor{bestbg}{RGB}{209,231,221}   

\begin{document}
\title{MedFM-Robust: Benchmarking Robustness of Medical Foundation Models}
%
\author{Xiangxiang Cui\inst{1}\textsuperscript{*} \and
Tianjin Huang\inst{2}\textsuperscript{*}
\and
Yifang Wang\inst{3}
\and
Lijie Hu\inst{4}
\and
Lu Yin\inst{5}\textsuperscript{$\dagger$}} 

\authorrunning{X. Cui, T. Huang, Y. Wang, L. Hu, and L. Yin}
%
\institute{Beijing Normal University, China \and
University of Exeter, United Kingdom
\and
University College London, United Kingdom
\and
Mohamed bin Zayed University of Artificial Intelligence, United Arab Emirates
\and
University of Surrey, United Kingdom\\
\email{l.yin@surrey.ac.uk}}



\maketitle              

\renewcommand{\thefootnote}{}
\footnotetext{\textsuperscript{*}These authors contributed equally to this work.}
\footnotetext{\textsuperscript{$\dagger$}Corresponding author.}
\renewcommand{\thefootnote}{\arabic{footnote}}

\begin{abstract}

Medical foundation models have achieved remarkable clinical performance, yet their robustness under
real-world perturbations remains underexplored. We present a robustness benchmark comprising 40 perturbation types (12 base, 28 medical-specific) across eight imaging modalities, evaluating five VLMs (LLaVA-Med, MedGemma, MedGemma-1.5, Gemini-2.5-flash and GPT-4o-mini) on VQA, visual grounding, and captioning, alongside two segmentation models (MedSAM, SAM-Med2D) with five fine-tuning strategies. Our findings reveal: (1) Fine-tuning strategy dominates robustness, with LoRA exhibiting nearly double the degradation of full fine-tuning, while SAM-Med2D's Adapter offers favorable efficiency-robustness trade-off. (2) Medical-specific perturbations disproportionately damage segmentation, with 9 of 15 top corruptions being domain-specific. (3) LoRA-tuned visual grounding drops over 40 points, whereas zero-shot captioning remains stable (<7\% drop). Zero-shot VQA shows model-dependent robustness—medical models drop under 20\% while Gemini-2.5-flash drops 54\%. General-purpose VLMs achieve higher VQA accuracy but fail on grounding; among medical VLMs, MedGemma demonstrates the best overall stability. These results provide deployment guidelines and underscore the necessity of domain-specific robustness evaluation for medical AI. Our code is available at: \url{https://abnerai.github.io/MedFM-Robust}.


\end{abstract}

\section{Introduction}


Medical foundation models (MedFMs) have emerged as transformative tools in healthcare, demonstrating capabilities across diverse clinical applications~\cite{2023Foundation,2023Large}. These models can be broadly categorized into two paradigms: Medical Vision-Language Models (Med-VLMs) and segmentation foundation models. Med-VLMs range from medical-specialized models such as LLaVA-Med~\cite{Li2023LLaVAMedTA} and MedGemma~\cite{Sellergren2025MedGemmaTR}, to general-purpose models like GPT-4o~\cite{hurst2024gpt} and Gemini~\cite{team2023gemini}, all capable of medical image understanding tasks including visual question answering (VQA), report generation, and visual grounding. Concurrently, the Segment Anything Model (SAM)~\cite{Kirillov2023SegmentA} has catalyzed a new generation of medical segmentation models, with adaptations like SAM-Med2D~\cite{Cheng2023SAMMed2D} and MedSAM~\cite{Ma2023SegmentAI}. The widespread clinical deployment of these models thus necessitates rigorous evaluation of their reliability under real-world conditions.

\begin{figure*}[t]
    \centering
    \includegraphics[width=\textwidth]{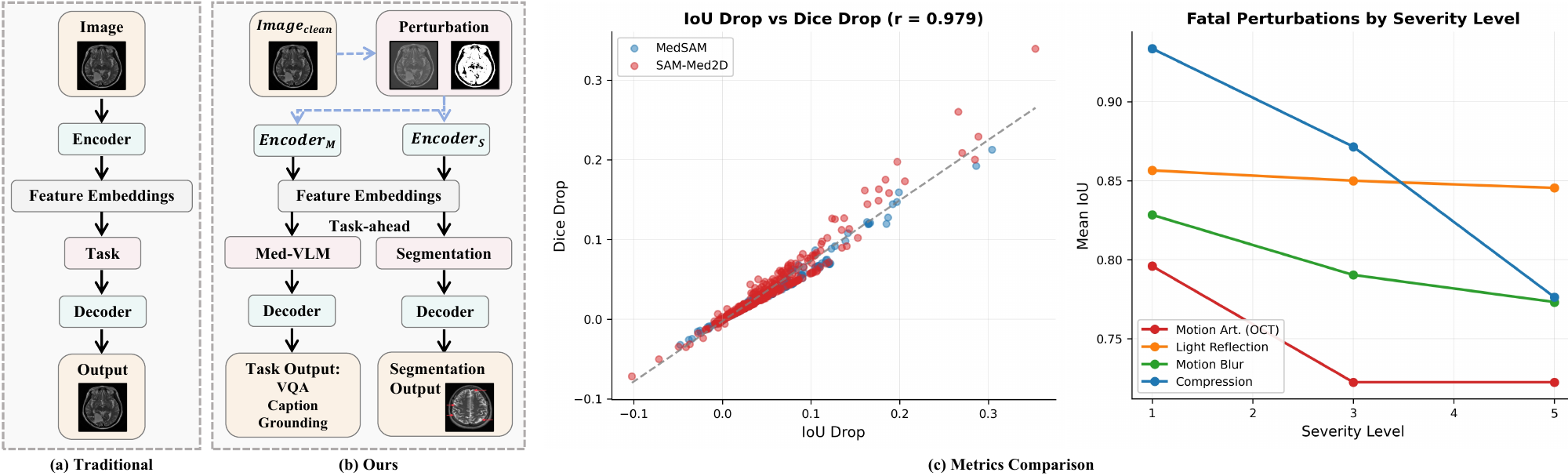}
    \caption{\textbf{Overview of our robustness evaluation and representative robustness drops.}
    \textbf{(a)} Traditional clean-image evaluation pipeline.
    \textbf{(b)} Our robustness benchmark applies modality-adaptive perturbations before the encoder and evaluates both Med-VLM tasks and segmentation under matched settings.
    \textbf{(c)} Metrics comparison: IoU drop closely tracks Dice drop, and representative fatal perturbations cause increasing performance degradation with higher severity levels.}
\end{figure*}

Despite impressive benchmark performance, medical foundation models face significant robustness challenges when deployed in clinical practice. Real-world medical images are inherently susceptible to various artifacts and perturbations arising from acquisition conditions, patient factors, and equipment variations~\cite{Suetens2017FundamentalsOM}. These include motion blur from patient movement, noise from low-dose protocols, and modality-specific degradations such as metal artifacts in CT, bias field inhomogeneity in MRI, speckle noise in ultrasound, and staining variations in pathology slides~\cite{Tellez2019QuantifyingTE} or or site-specific style shifts~\cite{zhao2026divide}. While extensive research has characterized model robustness in natural image domains through corruption benchmarks like ImageNet-C~\cite{DBLP:conf/iclr/HendrycksD19}, systematic evaluation of medical foundation model robustness remains critically understudied. Existing medical imaging benchmarks predominantly evaluate models on clean, curated datasets, creating a substantial gap between reported performance and real-world reliability.

Addressing this gap presents three fundamental challenges. First, medical imaging encompasses diverse modalities, each exhibiting distinct artifact patterns and degradation mechanisms. Generic perturbation models fail to capture modality-specific characteristics such as CT beam hardening, MRI ghosting artifacts, or histopathology stain variations~\cite{Reinhold2018EvaluatingTI}. Second, the field lacks a unified evaluation framework that comprehensively assesses robustness across both vision-language understanding tasks (VQA, captioning, grounding) and dense prediction tasks (segmentation), despite these capabilities often being deployed together in clinical systems. Third, the robustness implications of different fine-tuning strategies for medical foundation models remain largely unexplored, despite their widespread adoption in clinical adaptation scenarios.

In this work, we present a comprehensive framework for evaluating and enhancing the robustness of medical foundation models under domain-specific perturbations. Our contributions are threefold:




\begin{itemize}
\item We introduce a modality-adaptive perturbation pipeline spanning eight medical imaging modalities, with both base and modality-specific artifacts calibrated into five SSIM-guided severity levels for consistent degradation.

\item We establish a unified robustness benchmark for Med-VLMs and SAM-based segmentation models, evaluating VQA, captioning, and grounding across five VLMs (three medical-specialized, two general-purpose), and segmentation across five diverse clinical datasets covering dermoscopy, endoscopy, MRI, ultrasound, pathology, OCT, and CT.

\item We find that fine-tuning strategy critically determines robustness: LoRA exhibits the highest degradation across both VLMs and segmentation models, while full fine-tuning and adapter-based methods offer better robustness-efficiency trade-offs. General-purpose VLMs achieve strong zero-shot VQA but fail on grounding tasks requiring fine-tuning.
\end{itemize}
\section{Methods}

\begin{figure*}[t]
  \centering
  \includegraphics[width=\textwidth]{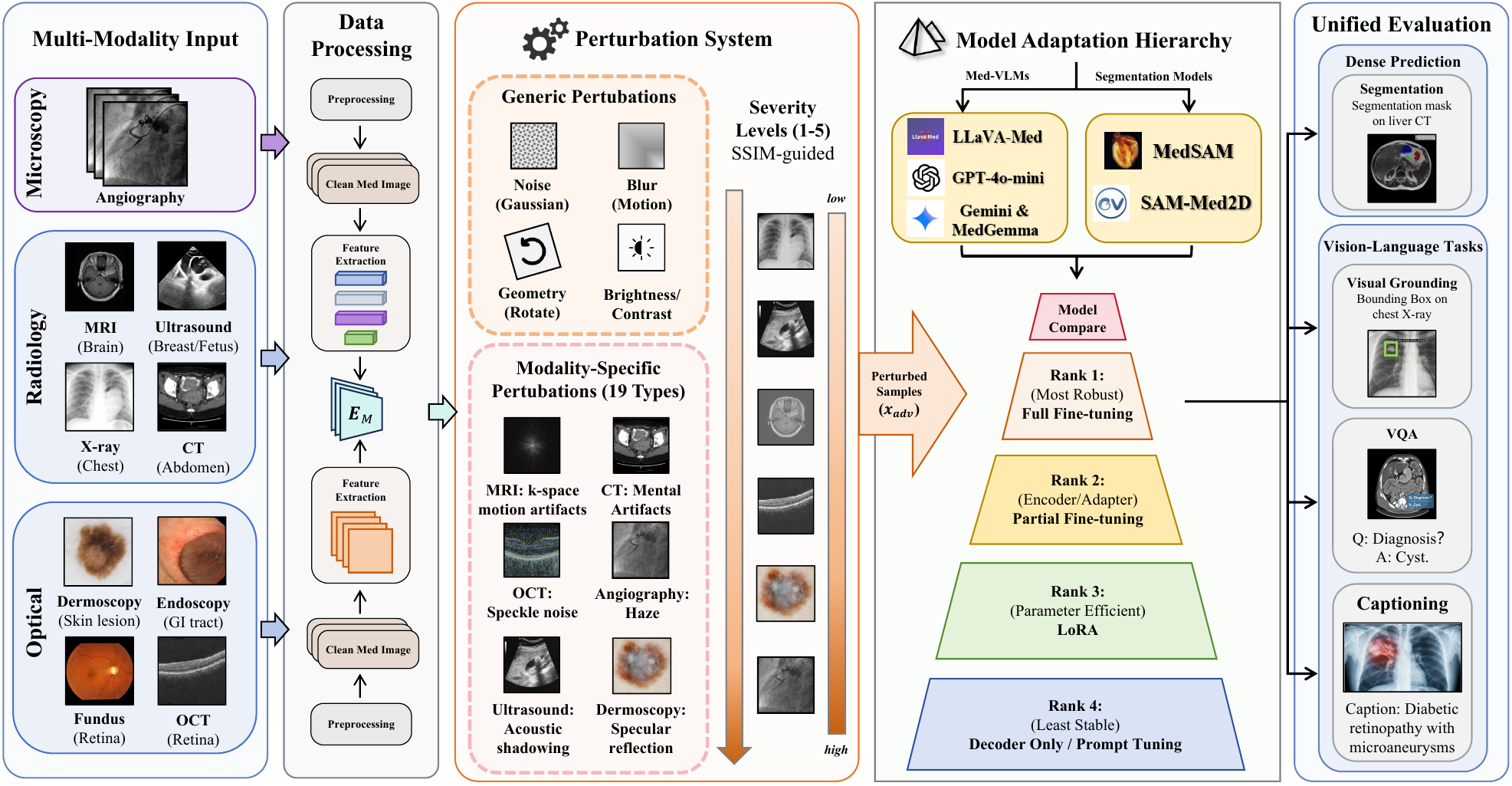}
  \caption{Overview of our robustness evaluation framework. We generate SSIM-calibrated perturbations across five severity levels, combining base corruptions with modality-specific artifacts. We benchmark three Med-VLMs and two SAM-based segmentation models under a unified protocol, and investigate multiple fine-tuning strategies across VQA, captioning, visual grounding, and segmentation tasks.}
  \label{fig:framework}
\end{figure*}

We present a comprehensive framework for evaluating the robustness of medical foundation models under realistic perturbations, integrating modality-adaptive perturbation generation with unified evaluation protocols for vision-language tasks and rigorous robustness assessment for segmentation models.

\subsection{Modality-Adaptive Perturbation Generation}

\subsubsection{Base Perturbations.}

We implement 12 base perturbation types applicable across all imaging modalities, categorized into three groups: \textit{noise} (Gaussian, salt-and-pepper, speckle), \textit{degradation} (Gaussian blur, motion blur, brightness, contrast, JPEG compression, pixelation), and \textit{geometric} (rotation, scaling, translation), enabling consistent cross-modality robustness evaluation.



\subsubsection{Modality-Specific Perturbations.}

zhBeyond base perturbations, we design modality-specific perturbations that simulate clinical artifacts across eight imaging modalities. For CT, we model metal-induced streaks and beam-hardening cupping. For MRI, we simulate bias-field inhomogeneity and ghosting. For ultrasound, we generate acoustic shadowing and reverberation. For pathology, we apply stain variations in HSV space. For endoscopy, we add specular reflections and bubbles. For OCT, we simulate shadow, blink, and defocus. For X-ray, we model scatter, exposure variation, and grid patterns. In task-specific experiments, we apply only the perturbations relevant to the modality of each dataset.

\subsubsection{SSIM-Guided Severity Calibration.}

We employ SSIM~\cite{Wang2004ImageQA} to ensure consistent degradation across five severity levels: Level 1 (SSIM 0.90--0.98), Level 2 (0.80--0.89), Level 3 (0.70--0.79), Level 4 (0.60--0.69), and Level 5 (0.50--0.59). For each perturbation type, we use binary search to find parameters achieving the target SSIM range, caching parameters for efficiency. However, not every perturbation reaches the most severe level, so the maximum number of iterations will be set to avoid excessive search time.

\subsection{Vision-Language Model Evaluation}

\subsubsection{Models and Datasets.}

We evaluate five foundation models: LLaVA-Med~\cite{Li2023LLaVAMedTA}, MedGemma~\cite{Sellergren2025MedGemmaTR}, MedGemma-1.5, GPT-4o-mini~\cite{2023GPT4VisionSC} and Gemini-2.5-flash~\cite{team2023gemini}. Evaluation spans three tasks: VQA on OmniMedVQA~\cite{Hu2024OmniMedVQAAN} (accuracy), captioning on ROCOv2~\cite{Pelka2018RadiologyOI} (BLEU, ROUGE-L, CIDEr), and visual grounding on MeCoVQA~\cite{Huang2024TowardsAM} (IoU@0.5). Among the three tasks, 500 samples were extracted from the dataset for evaluation.
For \textbf{visual grounding}, we employ LoRA~\cite{DBLP:conf/iclr/HuSWALWWC22} with rank $r=16$ and $\alpha=32$ on the vision encoder attention layers. Let $\mathcal{T}_{\text{response}}$ denote the bounding-box coordinate tokens. Training uses response-focused loss that only backpropagates through grounding output tokens:
\begin{equation}
    \mathcal{L}_{\text{GND}} = -\frac{1}{|\mathcal{T}_{\text{response}}|}
    \sum_{t \in \mathcal{T}_{\text{response}}}
    \log p_\theta(y_t \mid y_{<t}, I, Q),
\end{equation}
where $p_\theta$ is the model distribution parameterised by LoRA weights $\theta$,
$y_{<t}$ denotes all preceding tokens, $I$ is the input image, and $Q$ is the
query prompt. Gradients are backpropagated only through $\mathcal{T}_{\text{response}}$,
preventing instruction tokens from dominating the grounding supervision signal.

\subsection{Segmentation Model Robustness Evaluation}

\subsubsection{Models, Datasets, and Fine-tuning Strategies.}

We evaluate two SAM-based segmentation foundation models, SAM-Med2D~\cite{Cheng2023SAMMed2D} and MedSAM~\cite{Ma2023SegmentAI}, on five datasets spanning diverse clinical scenarios: ISIC 2016~\cite{Gutman2016SkinLA} (900 samples, dermoscopy), Kvasir-SEG~\cite{DBLP:conf/mmm/JhaSRHLJJ20} (1000 samples, endoscopy), Brain Tumor (3064 samples, MRI), and Glaucoma (5977 samples, Disc \& Cup). To study robustness enhancement under realistic perturbations, we further investigate five fine-tuning strategies: (1) decoder-only tuning, (2) encoder-only tuning, (3) full encoder tuning, (4) low-rank adaptation (LoRA), and (5) SAM-Med2D adapter tuning.

\subsection{Evaluation Protocol}

\subsubsection{Task-Specific Metrics.}
For segmentation, we use IoU and Dice coefficient:
\begin{equation}
\mathrm{IoU}_{\text{seg}}(P_m, G_m) = \frac{|P_m \cap G_m|}{|P_m \cup G_m|}, \qquad
\mathrm{Dice}_{\text{seg}}(P_m, G_m) = \frac{2|P_m \cap G_m|}{|P_m| + |G_m|},
\end{equation}
where $P_m \subseteq \Omega$ and $G_m \subseteq \Omega$ denote the predicted and
ground-truth masks for modality $m$ over pixel domain $\Omega$.
For VQA, we measure accuracy as
$\mathrm{Acc}=\frac{1}{N}\sum_{i=1}^{N}\mathbf{1}[\hat{a}_i=a_i^*]$.
For visual grounding, we report Acc@IoU${\geq}0.5$: a prediction is correct iff
the box-level overlap $|\hat{b}_i \cap b_i^*|/|\hat{b}_i \cup b_i^*| \geq 0.5$.
For captioning, BLEU~\cite{papineni2002bleu} measures clipped $n$-gram precision
with a brevity penalty, and CIDEr~\cite{vedantam2015cider} re-weights $n$-grams
by TF-IDF to reward clinically informative phrasing.

\subsubsection{Robustness Metric.}
To quantify performance degradation under perturbations, we report the absolute
performance drop. Let $\mathcal{P}_\tau$ denote the set of perturbation types
in category $\tau \in \{\text{base},\,\text{med-specific}\}$, $s \in \{1,\ldots,5\}$
the SSIM-calibrated severity level, and $M$ the task-specific metric
($\mathrm{IoU}_{\text{seg}}$ for segmentation, $\mathrm{Acc}_{\text{VQA}}$ and
$\mathrm{Acc}_{\text{GND}}$ for VQA and visual grounding, BLEU for captioning).
Lower values indicate better robustness. For aggregated analysis, we compute the
mean drop across all perturbation types within each severity level or perturbation category:
\begin{equation}
  \Delta_\tau^{(s)} = \frac{1}{|\mathcal{P}_\tau|}
  \sum_{p \in \mathcal{P}_\tau}
  \left( M_{\text{clean}} - M_{\text{perturb}}^{(p,s)} \right).
\end{equation}
\begin{figure}[ht]
    \centering
    \includegraphics[width=\linewidth]{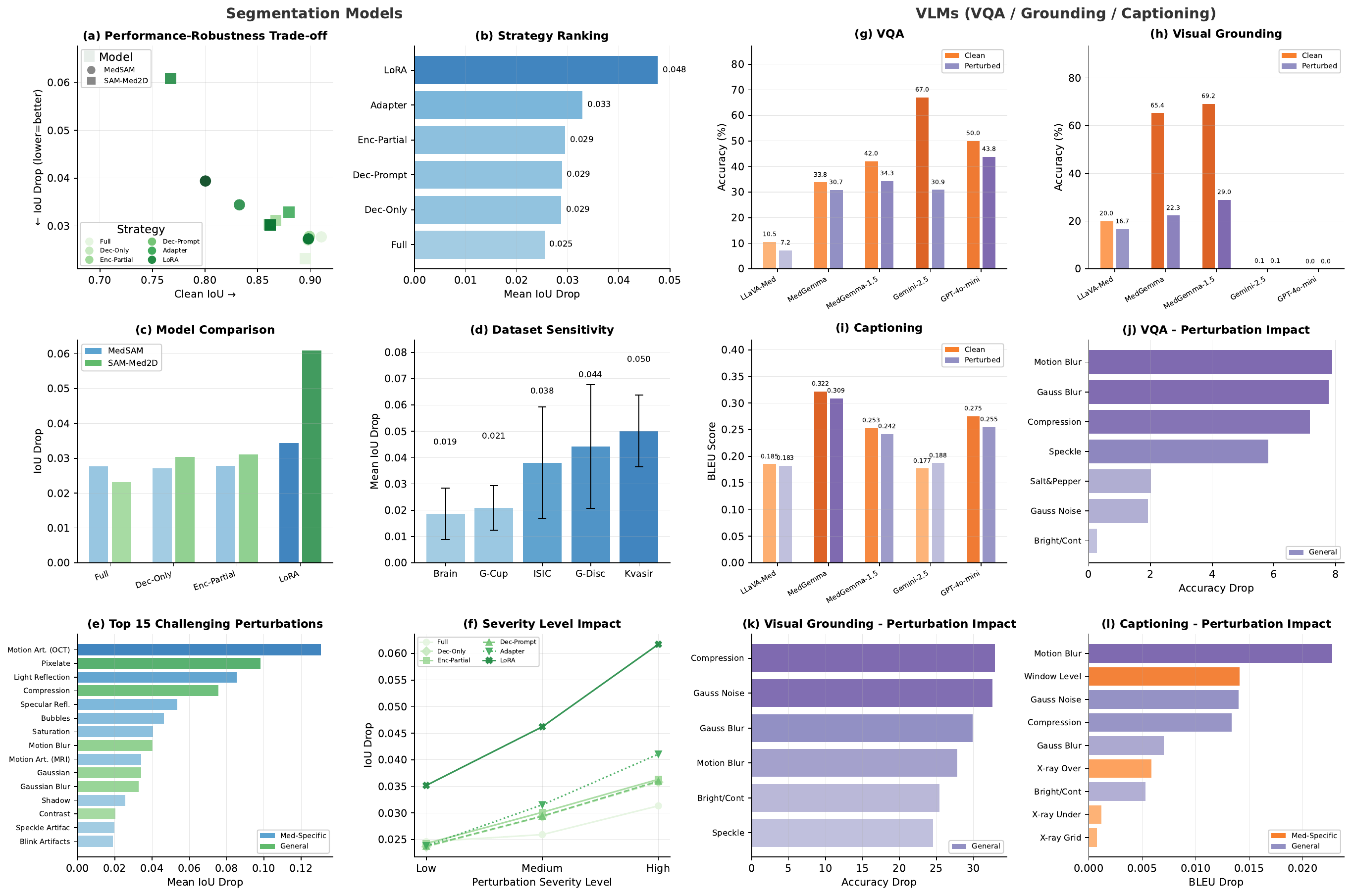}
    \caption{Comprehensive robustness evaluation of medical image segmentation models and VLMs under perturbations. \textbf{Left (Segmentation)}: (a) Performance-robustness trade-off. (b) Strategy ranking. (c) Model comparison. (d) Dataset sensitivity. (e) Top 15 perturbation types. (f) Severity level impact. \textbf{Right (VLMs)}: (g-i) Clean vs. perturbed performance on VQA, Grounding, and Captioning. (j-l) Perturbation impact.}
    \label{fig:seg_vlm_results}
\end{figure}


\section{Experiment}

\subsection{Experiment Setup}
All experiments use PyTorch on NVIDIA A100 GPUs. Segmentation fine-tuning uses AdamW with learning rate $10^{-4}$, weight decay 0.01, batch size 32, and 50 epochs with cosine scheduling. We evaluate on clean and perturbed images across five severity levels and report robustness using the absolute performance drop defined earlier.

\subsection{Qualitative Results in Segmentation} 

We evaluate five fine-tuning strategies on two SAM-based medical foundation models (MedSAM and SAM-Med2D) across five segmentation datasets, examining both clean performance and robustness under 19 perturbation types, \textit{e.g.}, Gaussian noise, motion blur, and modality-specific artifacts.


\noindent\textbf{Performance-Robustness Trade-off.} Fig.~\ref{fig:seg_vlm_results}(a) reveals a clear trade-off between clean segmentation accuracy and robustness to perturbations. Full fine-tuning achieves the highest clean IoU (0.89--0.91) while maintaining competitive robustness, positioning it in the desirable bottom-right region. In contrast, LoRA exhibits the largest performance degradation under perturbations despite reasonable clean performance, indicating that parameter-efficient methods may sacrifice robustness for efficiency.

\noindent\textbf{Strategy Ranking.} As shown in Fig.~\ref{fig:seg_vlm_results}(b), Full fine-tuning demonstrates the best overall robustness with a mean IoU drop of 0.025, followed by Dec-Only (0.029) and Enc-Only
(0.029). LoRA consistently ranks worst with a mean IoU drop of 0.048---nearly double that of Full fine-tuning. This ranking holds across both MedSAM and SAM-Med2D (Fig.~\ref{fig:seg_vlm_results}(c)), though SAM-Med2D generally exhibits higher sensitivity to perturbations than MedSAM across all strategies.

\begin{table*}[t]
\centering
\small
\caption{%
  Per-dataset robustness summary for \textbf{MedSAM} (top block) and
  \textbf{SAM-Med2D} (bottom block) across five benchmarks.
  \textit{Cl.}: clean IoU.
  $\Delta$B$\downarrow$: mean IoU drop under base corruptions.
  $\Delta$M$\downarrow$: medical-specific corruptions (lower = more robust).
}
\label{tab:robustness_merged}
\setlength{\tabcolsep}{4.5pt}
\renewcommand{\arraystretch}{1.15}

\resizebox{\textwidth}{!}{\begin{tabular}{>{\centering\arraybackslash}p{0.85cm} | c | c | ccc | ccc | ccc | ccc | ccc}
\toprule
\multirow{2}{*}{\textbf{Group}} &
\multirow{2}{*}{\textbf{Strategy}} &
\multirow{2}{*}{\textbf{Avg.~$\Delta$B$\downarrow$}} &
\multicolumn{3}{c|}{\textbf{ISIC 2016}} &
\multicolumn{3}{c|}{\textbf{Brain Tumor}} &
\multicolumn{3}{c|}{\textbf{Glaucoma Disc}} &
\multicolumn{3}{c|}{\textbf{Glaucoma Cup}} &
\multicolumn{3}{c}{\textbf{Kvasir-SEG}} \\

\cmidrule(lr){4-6}\cmidrule(lr){7-9}\cmidrule(lr){10-12}
\cmidrule(lr){13-15}\cmidrule(lr){16-18}

& & &
\textbf{Cl.} & \textbf{$\Delta$B$\downarrow$} & \textbf{$\Delta$M$\downarrow$} &
\textbf{Cl.} & \textbf{$\Delta$B$\downarrow$} & \textbf{$\Delta$M$\downarrow$} &
\textbf{Cl.} & \textbf{$\Delta$B$\downarrow$} & \textbf{$\Delta$M$\downarrow$} &
\textbf{Cl.} & \textbf{$\Delta$B$\downarrow$} & \textbf{$\Delta$M$\downarrow$} &
\textbf{Cl.} & \textbf{$\Delta$B$\downarrow$} & \textbf{$\Delta$M$\downarrow$} \\

\midrule
\rowcolor{secbg}
\multicolumn{18}{c}{\textbf{MedSAM}} \\
\midrule

\multirow{2}{*}{\rotatebox[origin=c]{90}{\bfseries\scriptsize Full}}
& \bestcell\textbf{Full}
  & \bestcell\gainval{.021}
  & \bestcell\textbf{.953} & \bestcell\gainval{.020} & \bestcell .034
  & \bestcell\textbf{.878} & \bestcell\underline{.022} & \bestcell\gainval{.011}
  & \bestcell\textbf{.952} & \bestcell\gainval{.011} & \bestcell\gainval{.023}
  & \bestcell\textbf{.907} & \bestcell\gainval{.007} & \bestcell .030
  & \bestcell\textbf{.862} & \bestcell .046 & \bestcell\gainval{.031} \\
& Enc-Only
  & \underline{.022}
  & \underline{.950} & \underline{.020} & \underline{.028}
  & \underline{.864} & \gainval{.013} & \gainval{.007}
  & \underline{.947} & .016 & .035
  & \underline{.892} & \gainval{.007} & .033
  & \underline{.843} & \underline{.041} & .036 \\
\cmidrule(l){2-18}

\multirow{2}{*}{\rotatebox[origin=c]{90}{\bfseries\scriptsize PEFT}}
& Dec-Prompt
  & \underline{.022}
  & \underline{.950} & \underline{.020} & \underline{.028}
  & \underline{.864} & .014 & .008
  & .946 & .016 & .035
  & .891 & \gainval{.007} & \underline{.031}
  & .842 & \gainval{.039} & .035 \\
& \textit{LoRA}
  & .025
  & .926 & \gainval{.015} & \gainval{.027}
  & .762 & .014 & .020
  & .919 & \loss{.048} & \loss{.084}
  & .761 & \loss{.021} & \underline{.024}
  & .795 & \gainval{.030} & \underline{.028} \\
\cmidrule(l){2-18}

\multirow{1}{*}{\rotatebox[origin=c]{90}{\bfseries\scriptsize Dec}}
& Dec-Only
  & \underline{.022}
  & .949 & \gainval{.020} & \gainval{.027}
  & .863 & \gainval{.013} & \gainval{.007}
  & \underline{.947} & .016 & .035
  & .890 & \underline{.006} & \gainval{.030}
  & .841 & \gainval{.039} & .035 \\

\midrule
\rowcolor{secbg}
\multicolumn{18}{c}{\textbf{SAM-Med2D}} \\
\midrule

\multirow{2}{*}{\rotatebox[origin=c]{90}{\bfseries\scriptsize Full}}
& \bestcell\textbf{Full}
  & \bestcell\gainval{.019}
  & \bestcell\textbf{.944} & \bestcell\underline{.029} & \bestcell\underline{.033}
  & \bestcell\textbf{.859} & \bestcell\underline{.024} & \bestcell .017
  & \bestcell\textbf{.950} & \bestcell\gainval{.010} & \bestcell\gainval{.018}
  & \bestcell\textbf{.899} & \bestcell\gainval{.002} & \bestcell\gainval{.007}
  & \bestcell\textbf{.826} & \bestcell .042 & \bestcell\gainval{.023} \\
& Enc-Only
  & .035
  & .932 & \loss{.049} & .035
  & \underline{.839} & \underline{.025} & \gainval{.005}
  & .931 & .037 & .044
  & .864 & .009 & .020
  & .772 & \underline{.041} & .031 \\
\cmidrule(l){2-18}

\multirow{2}{*}{\rotatebox[origin=c]{90}{\bfseries\scriptsize PEFT}}
& \underline{Adapter}
  & \underline{.029}
  & \underline{.935} & \underline{.043} & \gainval{.034}
  & \underline{.839} & .027 & \underline{.008}
  & \underline{.940} & \underline{.027} & .043
  & \underline{.884} & .011 & .022
  & \underline{.801} & \loss{.062} & .036 \\
& \textit{LoRA}
  & .051
  & .921 & \loss{.106} & .045
  & .730 & \loss{.045} & .016
  & .844 & \loss{.063} & \loss{.109}
  & .713 & \underline{.001} & \loss{.056}
  & .628 & .048 & \loss{.059} \\
\cmidrule(l){2-18}

\multirow{2}{*}{\rotatebox[origin=c]{90}{\bfseries\scriptsize Dec}}
& Dec-Prompt
  & .032
  & .929 & .053 & .036
  & .838 & \underline{.025} & \gainval{.004}
  & .927 & .037 & \underline{.041}
  & .855 & .008 & \underline{.017}
  & .760 & \gainval{.036} & \underline{.029} \\
& Dec-Only
  & .032
  & .929 & .054 & .036
  & .836 & \gainval{.024} & \gainval{.004}
  & .927 & .037 & \underline{.041}
  & .854 & \underline{.007} & \underline{.017}
  & .758 & \gainval{.036} & .031 \\

\bottomrule
\end{tabular}}
\end{table*}

\noindent\textbf{Dataset-Specific Sensitivity.} Robustness varies markedly across modalities and datasets. Brain MRI segmentation is the most stable (IoU drop: 0.019), likely benefiting from a more controlled acquisition environment. In contrast, Kvasir endoscopy exhibits the highest sensitivity (IoU drop: 0.050), consistent with the challenging and variable nature of gastrointestinal imaging. Tab.~\ref{tab:robustness_merged} reports the per-dataset results and confirms that full fine-tuning achieves the best clean performance while maintaining robustness across all datasets.

\noindent\textbf{Perturbation Analysis.} Failure cases are dominated by modality-specific corruptions. Motion Artifacts (OCT) and Light Reflection induce the performance drops, underscoring the need for domain-specific robustness evaluation. Among general perturbations, Pixelate is the most damaging. Notably, 9 of the top 15 most challenging perturbations are medical-specific, suggesting that standard robustness benchmarks may underestimate deployment risks.

\noindent\textbf{Severity Level Impact.} Performance degrades monotonically as perturbation severity increases, but the degradation rate differs across strategies. LoRA exhibits the steepest curve, with IoU drop rising from 0.028 at low severity to 0.065 at high severity. Full fine-tuning shows the flattest curve, suggesting that updating all parameters helps learn more robust feature representations. The gap between strategies widens at higher severity levels, indicating that robustness differences become more pronounced under severe distribution shifts.


\subsection{Vision-Language Tasks Evaluation}

We evaluate five medical vision-language models (LLaVA-Med, MedGemma, MedGemma-1.5, GPT-4o-mini and Gemini-2.5-flash) on VQA, Visual Grounding, and Captioning. VQA and Captioning are evaluated in a zero-shot setting, while Visual Grounding uses LoRA fine-tuning.

\noindent\textbf{Task-Specific Robustness.} Robustness differs sharply across tasks and model types. Visual Grounding, which requires LoRA fine-tuning, degrades severely for medical models: MedGemma drops from 65.4\% to 22.3\% and MedGemma-1.5 from 69.2\% to 29.0\%. General-purpose models evaluated zero-shot fail entirely on this task, indicating that Grounding inherently requires task-specific adaptation \textbf{for precise localization}. For zero-shot VQA, Gemini-2.5 achieves the highest clean accuracy (67.0\%) but suffers the largest drop (36.1 points, 54\% relative), while GPT-4o-mini and medical models behave more stably with drops under 8 points. Captioning remains highly robust across all models with drops below 0.02 BLEU \textbf{even at high severity}. This pattern suggests that fine-tuning boosts task-specific performance at the expense of robustness, whereas zero-shot inference preserves pretrained stability (see Fig.~\ref{fig:seg_vlm_results}(g)--(i)).

\noindent\textbf{Perturbation Impact.} Perturbations affect tasks in different ways. For LoRA-fine-tuned Grounding, Compression Artifacts (32.9 drop) and Gaussian Noise (32.6 drop) are most damaging, mirroring the sensitivity of fine-tuned segmentation models. For zero-shot VQA, Motion Blur and Gaussian Blur cause the largest drops (around 8 points), while medical-specific perturbations such as Window Level and X-ray artifacts have moderate impact. Captioning remains resilient, with all perturbations causing drops below 0.025 BLEU, suggesting that caption generation relies on higher-level semantic cues that are less sensitive to low-level corruptions (see Fig.~\ref{fig:seg_vlm_results}(j)--(l)).

\noindent\textbf{Model Comparison.} General-purpose and medical-specialized models exhibit clear trade-offs. In zero-shot VQA, Gemini-2.5 reaches the highest accuracy (67.0\%) but the weakest robustness (54\% relative drop), whereas GPT-4o-mini is more balanced (50.0\% clean, 12\% drop) and performs strongly on captioning. Medical models are more stable, with MedGemma showing the smallest drops on VQA (3.1) and captioning (0.013). For grounding, general-purpose models fail in the zero-shot setting (0--10\%), while LoRA-fine-tuned medical models achieve high performance (MedGemma-1.5: 69.2\%) at the cost of robustness. This matches the segmentation results: fine-tuning enables specialized capabilities but increases sensitivity to perturbations (see Fig.~\ref{fig:seg_vlm_results}(a)-(c)).

\section{Conclusions}

We present a robustness benchmark for medical foundation models with 40 perturbations (12 base, 28 modality-specific) across eight modalities, evaluating segmentation (MedSAM, SAM-Med2D) and VLMs (LLaVA-Med, MedGemma, MedGemma-1.5, Gemini-2.5-flash, GPT-4o-mini). Robustness is mainly determined by fine-tuning strategy: LoRA degrades nearly twice as much as full fine-tuning. Modality-specific corruptions dominate segmentation failures top 9 corruptions. Task formulation further matters: LoRA-tuned Grounding drops >60\%, zero-shot Captioning stays <7\%, and VQA robustness is model-dependent (medical <20\% vs.\ Gemini-2.5-flash 54\%). For deployment, full fine-tuning is most robust, with SAM-Med2D adapters as a lightweight alternative. General-purpose VLMs excel at zero-shot VQA but fail on Grounding, while MedGemma is the most consistently robust medical VLM.
\bibliographystyle{splncs04}
\bibliography{mybibliography}
\end{document}